\documentclass[onecolumn,hf]{ceurart}

\usepackage{adjustbox}
\usepackage{appendix}
\usepackage{array}
\usepackage{booktabs}
\usepackage{colortbl}
\usepackage{csquotes}
\usepackage{cuted}
\usepackage[shortlabels]{enumitem}
\usepackage{graphicx}
\usepackage{multirow}
\usepackage{subcaption}
\usepackage{tabularx}
\usepackage{tcolorbox}
\usepackage{ulem}
\usepackage{xcolor}
\usepackage{xtab}
\usepackage{xurl}

\tcbset{before upper={\parindent0em}}
\newtcolorbox[auto counter]{quotebox}{boxsep=-5pt,colback=gray!20,width=1.05\columnwidth}

\newcounter{myquote}
\renewcommand{\themyquote}{\ref{sec:appendix:quotes}\arabic{myquote}}
\makeatletter

\newcolumntype{s}{>{\columncolor[rgb]{0.9,0.9,0.9}}c}

\AtBeginDocument{%
  \providecommand\BibTeX{{%
    \normalfont B\kern-0.5em{\scshape i\kern-0.25em b}\kern-0.8em\TeX}}}

\begin{document}
\copyrightyear{2022}
\copyrightclause{Copyright for this paper by its authors. Use permitted under Creative Commons License Attribution 4.0 International (CC BY 4.0).}

\conference{ATT '22: Workshop on Agents in Traffic and Transportation, July 25, 2022, Vienna, Austria}

\title{The Real Deal: A Review of Challenges and Opportunities in Moving Reinforcement Learning-Based Traffic Signal Control Systems Towards Reality}

\author{Rex Chen}[email=rexc@cmu.edu]
\author{Fei Fang}[email=feifang@cmu.edu]
\author{Norman Sadeh}[email=sadeh@cs.cmu.edu]
\address{Institute of Software Research, School of Computer Science, Carnegie Mellon University}

\begin{abstract}
Traffic signal control (TSC) is a high-stakes domain that is growing in importance as traffic volume grows globally. An increasing number of works are applying reinforcement learning (RL) to TSC; RL can draw on an abundance of traffic data to improve signalling efficiency. However, RL-based signal controllers have never been deployed. In this work, we provide the first review of challenges that must be addressed before RL can be deployed for TSC. We focus on four challenges involving (1) uncertainty in detection, (2) reliability of communications, (3) compliance and interpretability, and (4) heterogeneous road users. We show that the literature on RL-based TSC has made some progress towards addressing each challenge. However, more work should take a systems thinking approach that considers the impacts of other pipeline components on RL.
\end{abstract}
 
\maketitle

\section{Introduction}
\label{sec:intro}

As the traffic volume of metropolitan areas continues to grow worldwide, gridlock is becoming an increasingly prevalent concern. According to the \textit{2021 Urban Mobility Report} \cite{Schrank2021}, gridlock led to over 4 billion hours in travel delay and \$100+ million in congestion costs across the United States in 2021. This not only impacts commercial productivity but also has environmental consequences. One important mechanism for alleviating gridlock is improving the timing of traffic signals \cite{Chin2004}. Historically, most jurisdictions have used fixed timing plans based on traffic models, which assume fixed values of factors such as lane volumes and arrival rates \cite{Zheng2019a}. To minimize implementation burden, traditional \emph{traffic signal control} (TSC) either uses one fixed plan throughout the entire day, or rotates through several plans depending on the time of the day. However, fixed plans cannot respond in real time to changes in traffic demand \cite{Zheng2019a,Greguric2020}.

Large traffic volumes also offer an abundance of data that can be used for real-time optimization of signal timing plans. Many deployed systems combine logic-triggered state changes with data-driven searches over sets of schedules \cite{Zheng2019a}. However, an increasing number of approaches traverse larger search spaces using optimization and scheduling algorithms \cite{Smith2013}. Among these approaches, \emph{reinforcement learning} (RL) has yielded significant improvements over fixed and actuated TSC algorithms in simulations \cite{Chen2020}. RL allows systems to learn from the consequences of their decisions, which enables them to achieve continuous self-improvement. Deployments of RL algorithms have achieved success in a variety of complex domains involving human interaction, such as card games \cite{Brown2017}, real-time strategy games \cite{Vinyals2019}, and other applications in transportation such as dispatching for ride-hailing services \cite{Qin2020}.

However, to our knowledge, RL-based TSC algorithms have never been deployed. This is in spite of the fact that papers introducing novel algorithms in this area commonly list real-world deployment as a goal for future work \cite{Noaeen2022}. We believe that this discrepancy has arisen due to a focus on methodological contributions, instead of on a holistic systems thinking approach based on the data-to-deployment pipeline \cite{Perrault2019}. If RL-based signal controllers are to achieve success in deployment, domain experts in TSC and in RL must have a shared view of the problem. We take a step towards bridging the gap between research and deployment by providing the first review of challenges that may arise from end-to-end deployments of RL-based TSC, which we intend to provide a common basis of collaboration between research in TSC and RL.

We begin by describing our review methodology in Section \ref{sec:intro:method}. Then, we provide a high-level review of the fields of TSC and RL in Section \ref{sec:related}, followed by a more detailed problem formulation for RL-based TSC in Section \ref{sec:problem}. Next, we explore four engineering challenges. For each of these challenges, we will provide a review of (1) how these challenges are significant concerns for the state of the art in RL-based TSC; (2) what practical considerations relevant to these challenges have arisen in deployments of non-RL TSC systems; and (3) what progress has been made in the RL-based TSC literature towards solving these challenges.

\begin{itemize}
    \item \textbf{Uncertainty in detection.} (Section \ref{sec:detect}) Typically, RL-based TSC algorithms learn based on metrics such as queue length or travel time. These require accurate vehicle detection technologies, which may not always be available in the field. Strategies to deal with detector uncertainty and failure are a prerequisite of deployment.
    
    \item \textbf{Reliability of communications.} (Section \ref{sec:comm}) Some decentralization is necessary for RL-based TSC. Coordination between intersections is important for optimizing network-level metrics, yet most work in RL-based TSC has not considered the practicalities of dealing with failure and latency in inter-intersection communications.
    
    \item \textbf{Compliance and interpretability.} (Section \ref{sec:interp}) Jurisdictions will not have confidence in RL-based signal controllers without assurances about compliance to standards (e.g., minimum green time) and safety requirements. The interpretability of models is important for ensuring that signalling plans can be audited and adjusted by stakeholders.
    
    \item \textbf{Heterogeneous road users.} (Section \ref{sec:roadusers}) Most simulations for RL-based TSC assume that all cars are the same size and have the same free-flow speed. However, cars share the road with pedestrians, buses, emergency vehicles, and other road users. Algorithms must detect and respond to the needs of different road users in a safe, equitable manner.
\end{itemize}

Finally, we end with concluding thoughts and suggestions for future work in Section \ref{sec:conclusion}.

\subsection{Methodology}
\label{sec:intro:method}
To obtain an overview of the domain of RL-based TSC, we conducted a targeted search on Google Scholar with the keywords ``traffic signal''/``traffic light'', ``reinforcement learning'', and ``review''/``survey''. We identified the four challenges addressed in the following sections through these reviews. From here, we conducted snowball sampling based on their citations to locate papers in the RL literature that discuss these challenges. For RL papers, we focused on those published after 2015, since this field has rapidly evolved over the past several years. We also performed additional targeted Google Scholar searches to find literature which describes non-RL deployments of TSC, by searching the keywords ``traffic signal''/``traffic light'' and ``adaptive'' in conjunction with the following keywords:
\begin{itemize}
    \item For \textbf{Section \ref{sec:detect}}, ``uncertainty'', ``noise'', ``sensing error'', ``accuracy'.
    
    \item For \textbf{Section \ref{sec:comm}}, ``coordination'', ``communication'', ``closed loop'', ``message'', ``NTCIP''.
    
    \item For \textbf{Section \ref{sec:interp}},  ``compliance'', ``safety'', ``accountability'', ``interpretability''/``explainability''.
    
    \item For \textbf{Section \ref{sec:roadusers}}, ``pedestrian''/``leading pedestrian interval'', ``cyclist'', ``transit'', ``emergency vehicle'', ``priority'', ``preempt''.
\end{itemize}

\section{Related work}
\label{sec:related}

\subsection{Traffic signal control}
\label{sec:related:tsc}
Traffic signal control (TSC) aims to allocate green time at an intersection to traffic moving in different directions. Every \emph{approach} (roadway entering the intersection) is split into lanes for forward, left-turn, and (possibly) right turn movements (which may be assumed to always be permissible) \cite{Gordon2005,Zheng2019b}. For efficiency, pairs of compatible movements are often arranged into \emph{phases} and signalled simultaneously \cite{Noaeen2022,Koonce2008,Wei2019a}. The task is to find some division of green time between phases for each intersection in a road network, which maximizes metrics such as the throughput of the network. We refer the reader to \cite{Eom2020} for details of the problem formulation.

Different approaches to dividing green time include choosing phase durations or phase sequences, or fixing a phase sequence within a \emph{cycle} and choosing the length of the cycle or the proportions of each phase within the cycle \cite{Noaeen2022,Gordon2005,Wei2019a}. Three main types of algorithmic approaches exist. In \emph{fixed-time control}, which has historically been a popular strategy \cite{Zheng2019a}, a small number of fixed plans are optimized based on past traffic data under the assumption of uniform demand. In \emph{actuated control}, detector inputs (such as vehicle presence data from loop detectors) are used in conjunction with a fixed set of logical rules. Finally, \emph{adaptive control} uses more complex prediction and optimization algorithms to control signalling plans \cite{Gordon2005,Eom2020}.

\subsection{Reinforcement learning}
\label{sec:related:rl}
One emerging approach to adaptive control has been reinforcement learning (RL). RL is a sequential decision-making paradigm wherein agents learn how to act through trial-and-error interactions with an environment. The goal of RL is to learn \emph{policies}, which describe how agents should act given the state of the environment. Early work in reinforcement learning during the 1980s and 1990s, which included the seminal $Q$-learning algorithm \cite{Watkins1992}, relied on tabular enumeration of environment states and agent actions. RL remained relatively difficult to scale until the emergence of methods based on function approximation in the 2010s, specifically the use of neural networks for \emph{deep RL} \cite{Mnih2015}. Since then, the popularity and complexity of RL has experienced explosive growth. Game-playing deep RL agents have achieved superhuman performance in card games and video games with high-dimensional state and action spaces and real-time decision-making, such as AlphaGo (Go) \cite{Silver2016}, Libratus (heads-up no-limit poker) \cite{Brown2017}, and AlphaStar (StarCraft II) \cite{Vinyals2019}. Deep RL has also found novel applications in practical domains such as robotics, natural language processing, finance, and healthcare \cite{Li2018b}. Transportation has been one of the most significant applications of deep RL, with tasks including autonomous driving \cite{Kiran2021}, vehicle dispatching \cite{Qin2020} and routing \cite{Nazari2018}, and traffic signal control (see Section \ref{sec:related:review}). We refer the reader to \cite{Sutton2018} for an in-depth review of the history of reinforcement learning.

The body of work that we review in this paper can be seen as a parallel to work in RL for robotics that attempts to close the gap between simulations and reality. RL methods, especially deep RL methods, require an abundance of data to learn from environmental interactions. Due to the cost of real-world data collection, simulators are often employed instead to generate large quantities of interactions. However, simulators can never perfectly emulate reality. This problem, which is referred to as the reality gap \cite{Mouret2017}, has been addressed by the \emph{sim-to-real} literature. Some sim-to-real methods employ randomization in sensors and controllers to learn robust policies (\emph{domain randomization}); some explicitly model the reality gap and try to unify the feature spaces of the source and target environments (\emph{domain adaptation}); some train policies to generalize across different tasks (\emph{meta-RL}); some attempt to learn from demonstrations of behaviour in target environments (\emph{imitation learning}); and others attempt to improve simulators. We refer the reader to \cite{Zhao2020,Dimitropoulos2022} for surveys of these methods. In this work, we draw parallels between some of these methods and developments in RL-based TSC. However, at the same time, TSC involves unique challenges that are usually not present in robotics. Environments in robotics where sim-to-real methods have been applied (see \cite{Zhao2020}) are usually highly controlled with well-defined objectives (e.g., \cite{Andrychowicz2020}) and minimal interaction with other agents. However, TSC may be affected by varying environmental conditions and large numbers of road users.

\subsection{Related reviews}
\label{sec:related:review}
Various reviews of applications of RL in TSC have been published. While each of the following reviews captures distinct aspects of the field that are highly relevant to our work, none of them have focused on the key issue of practical engineering challenges that present barriers to deployment, and --- crucially --- how to solve them instead of leaving them as open problems.

\cite{Abdulhai2003}, \cite{Bazzan2009}, and \cite{Bazzan2013} provided brief syntheses of early RL-based TSC methods in reviews of applications of AI in transportation. \cite{Mannion2016} and \cite{Yau2017} were the first to take a systematic approach to reviewing RL-based TSC algorithms; the former performed the first experimental comparison of RL algorithms with a synthetic network, while the latter addressed data sources such as models of road networks and vehicle arrivals. Both reviewed state, action, and reward formulations. These reviews considered traditional algorithms in RL such as Q-learning and SARSA.

With the increasing popularity of deep learning to address challenges of scalability in RL, \cite{Greguric2020,Rasheed2020} (the latter a follow-up to \cite{Yau2017}) both reviewed deep RL methods for TSC and provided recommendations for designing novel deep RL-based TSC algorithms. \cite{Greguric2020} focused on choosing state, action, and reward representations, with some discussion of data processing, but did not consider downstream challenges in deployment. \cite{Rasheed2020} provided a broad overview of various algorithm and architecture designs with less of a focus on practicalities.

Both \cite{Wei2019a,Wei2021} reviewed alternative state, action, and reward formulations among deep RL-based TSC algorithms, as well as options for inter-agent coordination and simulation-based evaluation. They outlined, but did not investigate, challenges to deployment. \cite{Wei2019a} further compared deep RL-based algorithms to traditional actuated and adaptive methods. Likewise, as part of a wider review on deep RL for intelligent traffic systems, \cite{Haydari2022} reviewed problem formulations and the history of algorithmic developments for RL-based TSC. Finally, \cite{Noaeen2022} performed a highly systematic overview of the past 26 years of research in this domain that provides quantitative support for some of the patterns that we identify.

\section{Reinforcement learning-based traffic signal control}
\label{sec:problem}
In this section, we establish the basic problem formulation for RL-based TSC. As noted in Section \ref{sec:related:review}, previous reviews have covered different design decisions for RL algorithms and training frameworks at length. We provide only the details that are most salient to the challenges that we discuss, and we refer the reader to \cite{Greguric2020,Wei2019a,Rasheed2020} for syntheses of how this general framework has been instantiated. The process involves three steps: (1) problem formulation (Section \ref{sec:problem:mdp}, \ref{sec:problem:extend}), (2) simulation (Section \ref{sec:problem:sim}), and (3) algorithm design (Section \ref{sec:problem:alg}).

\subsection{Markov decision processes}
\label{sec:problem:mdp}
The most common sequential decision-making problem formulation for RL-based TSC is the \emph{Markov decision process} (MDP), which can be formally described as a tuple $(S,A,T,R,\gamma)$. In an MDP, there is a single controller agent interacting with an environment, typically one intersection, over a number of time steps $t \in \{1..T\}$. At each time step $t$:

\paragraph{State} 
The controller receives a representation of the current intersection \emph{state}, $s_t \in S$, where $S$ is the set of all possible states. For reasons of tractability, $s_t$ is usually some kind of abstracted, numerical representation of the intersection. As reviewed by \cite{Noaeen2022}, five of the most commonly included elements of state representations at the intersection level are (1) the numbers of queueing vehicles in lanes (queue length, 38\%), (2) the current phase (11\%), (3) the number of vehicles (10\%), (4) the positions of vehicles (6\%), and (5) the speeds of vehicles (6\%).
    
\paragraph{Action} 
Based on $s_t$, the controller chooses an \emph{action} (i.e. signalling decision) $a_t \in A$ to take, where $A$ is the set of all possible actions (assumed to be the same for each state). For a majority (62\% per \cite{Noaeen2022}) of works, agents choose the next phase and phase duration. In other works (32\% per \cite{Noaeen2022}), the action space is based on cycles, including varying the phase split and sequence in fixed-length cycles or varying the cycle length.
    
\paragraph{State transitions} 
The action affects the intersection immediately through a probabilistic transition to the next state, which represents the immediate effect of the signalling decision. Given $s_t, a_t$, each possible next state $s_{t+1} \in S$ occurs with probability 
\begin{align*}
    T(s_t, a_t, s_{t+1}) = \Pr(s_{t+1} \mid s_t, a_t)
\end{align*}
where $\sum_{s_{t+1}} T(s_t, a_t, s_{t+1}) = 1, \forall s_t, a_t$. Most work in RL-based TSC adopts a model-free approach that does not model $T$ explicitly \cite{Wei2019a}, since the typical sizes of the state and action spaces make the explicit learning and representation of $T$ prohibitively costly.
    
\paragraph{Reward} 
After taking the action, the controller receives a numerical \emph{reward} $r_t(s_t, a_t, s_{t+1}) \in \mathbb{R}$, which is some measure of how good the signalling decision was. In TSC, this can be based on the updated state (per \cite{Noaeen2022}, 30\% use queue lengths; 6\% use vehicle counts), or on vehicle-specific quality metrics (per \cite{Noaeen2022}, 13\% use the delays of vehicles, in terms of increase in travel time; 9\% use the waiting times of vehicles; 4\% use the throughput of intersections). The agent uses rewards to learn` how good signalling decisions are in various intersection states.
    
\paragraph{Policies} 
The goal of the controller is to learn a policy $\pi: S \rightarrow A$ that maps the current state to the action that it should take: $a_t = \pi(s_t)$, which should be optimal in the sense that it maximizes cumulative rewards $\sum_{t=1}^T r_t$. 
    
As this is a sequential problem, the agent cannot greedily choose actions to maximize estimated rewards at every time step, because its choices may have ramifications for future time steps in terms of what the state will be (e.g., clearing one intersection at time step $t$ may cause congestion at another during time step $t + 1$). The \emph{$Q$-value function} $Q(s, a)$ encodes this notion by incorporating a decaying contribution from the expected rewards of future time steps:
\begin{align*}
    Q(s, a) = \sum_{s'} \Pr(s' \mid s, a) \left[r(s, a, s') + \gamma \max_{a'} Q(s', a')\right]
\end{align*}
where $\gamma \in [0,1]$ is a discount factor. Intuitively, the $Q$-value describes the value of making a signalling decision in a state, given the best decision is always made in the future. The optimal policy maximizes the $Q$-value:
\begin{align*}
    \pi^*(s_t) = \mathrm{argmax}_a Q(s_t, a)
\end{align*}

MDP-based formulations for RL-based TSC make several assumptions: (1) the intersection state is fully observable; (2) state transitions always follow the same probabilistic model $T$ given the current state $s_t$ and signalling decision $a_t$; and (3) rewards accumulate additively.

\subsection{Other frameworks}
\label{sec:problem:extend}
Several extensions of the MDP framework have been leveraged for applications in TSC. Each of these frameworks approaches more closely to reality than MDPs, yet incurs additional computational costs. Previous reviews have not addressed these frameworks in detail.

\paragraph{Partially observable MDPs}
In \emph{partially observable MDPs} (POMDPs), the agent does not directly observe the state $s_t$. Instead, there is a space of \emph{observations} $\Omega$, which is usually some partial representation of the intersection state in TSC. The controller's observations $o_t \in \Omega$ are assumed to be samples from some probability distribution $O(s_t, o_t) = \Pr(o_t \mid s_t)$, with $\sum_{o_t} \Pr(o_t \mid s_t) = 1, \forall s_t$. The controller never knows exactly what state it is in; instead, it maintains a probability distribution over states known as a belief state, $b_t \in \Delta^{|S|}$, which is based on an initial state distribution $b_0 \in \Delta^{|S|}$. Thus, a POMDP can be described using a tuple $(S, A, T, R, \Omega, O, b_0, \gamma)$.

POMDPs are useful for when the same state may be observed differently by agents depending on some randomness. Works in RL-based TSC that use a POMDP framework \cite{Mannion2016,Xie2020,Li2021a} either use observations to represent local intersection states in a road network (e.g., \cite{Xie2020}), or to represent incomplete inputs from detectors (as is done by \cite{Li2021a} for connected vehicle data).

\paragraph{Markov games}
Known either as the \emph{Markov game} or the stochastic game, this framework generalizes MDPs to the setting of \emph{multi-agent RL}. The key difference is that there is a set of agents $N = \{1..n\}$, and each agent has an action space $A_i$ with $A = \prod_i A_i$. At each time step, every agent $i$ takes its own action $a_t^{(i)}$ and its own reward $r_t^{(i)}(s_t, a_t, s_{t+1})$. Thus, a Markov game can be described using a tuple $(N, S, A, T, R, \gamma)$.

Markov games are useful for network-level TSC, where there are multiple intersections to consider. Works in RL-based TSC that use a Markov game framework \cite{Wei2019a,Wei2019c} generally define each intersection's controller as an agent. This is opposed to combining all state and action spaces into a single agent, which may be difficult to scale due to high dimensionality.

In this setting, the optimal policies for controllers will always depend on knowledge of the states and actions of other controllers in the road network. It is possible for all of the policies to be learned in a centralized fashion with a single algorithm instance, or in a fully decentralized fashion where the other controllers are viewed as part of the environment. Centralized training can be computationally intensive, but decentralized training may result in the suboptimality of policies. Mechanisms for coordination can help to bridge this gap \cite{Zhang2021b}.

\paragraph{Decentralized POMDPs}
Decentralized POMDPs (\emph{dec-POMDPs}), or partially observable Markov games, are a natural combination of POMDPs and Markov games. As applied to the TSC setting, we again assume that there are multiple intersection controllers $N$; in addition to the actions $a_t^{(i)}$ and rewards $r_t^{(i)}$, we also assume that each intersection controller receives its own observations $o_t^{(i)} \in \Omega_i$ with $\Omega = \prod_i \Omega_i$. Thus, a dec-POMDP can be described using a tuple $(N, S, A, T, R, \Omega, O, b_0, \gamma)$. 

Although MDPs are the dominant framework in RL-based TSC, a sizeable minority of work has used dec-POMDPs. Such work \cite{Alegre2021,Li2021b,Ma2020,Yang2019,Yang2021,Jiang2022} generally defines observations as the local state for the controller's intersection, as opposed to the state of the entire road network. However, dec-POMDPs are also computationally the most difficult to solve, and the design of efficient solution methods is an area of active research \cite{Amato2018}.

\subsection{Simulators}
\label{sec:problem:sim}
As noted in Section \ref{sec:problem:mdp}, it is infeasible to model the transition probabilities $T$ explicitly for TSC. This has motivated the use of \emph{traffic simulators} to generate training experiences. Non-RL methods in TSC have also made extensive use of simulators, which has resulted in many proprietary and open-source traffic simulators that can be used for RL applications \cite{Wei2019a}. 

Traffic simulators can be roughly subdivided into three groups based on the level of granularity. \emph{Microscopic} simulators provide the most granular simulations, as they simulate the behaviour of individual vehicles (including acceleration, deceleration, and lane changing). This behaviour is generally based on car-following models, which model the acceleration of a vehicle as a time series that depends on its own speed and other vehicles' speeds. \emph{Macroscopic} simulators, by contrast, aggregate vehicles into flows, with time series describing the volume, speed, and density of flows between different points in a road network. \emph{Mesoscopic} simulators provide an intermediate solution between microscopic and macroscopic simulators that balances detail and computational efficiency; they may either model individual vehicles as flows, or organize groups of vehicles into platoons. We refer the reader to \cite{Barcelo2010} for an overview of traffic models that have been used for these types of simulations. 

In RL-based TSC, the most popular traffic simulator (as reviewed by \cite{Noaeen2022}) is SUMO \cite{Krajzewicz2002}. SUMO is an open-source microscopic simulator that can flexibly simulate various types of signalling plans, road structures, and road user types. The next most popular simulators include proprietary microscopic/mesoscopic simulators (VISSIM, PARAMICS, AIMSUN) and the open-source microscopic simulator GLD \cite{Wiering2004}. Concerns about slowness have also led to the development of dedicated simulators for RL. For instance, CityFlow \cite{Zhang2019} simplifies simulations greatly in exchange for a 20-fold speedup over SUMO. As for the contents of simulations, a majority of works (62\% per \cite{Noaeen2022}) have used synthetically-generated road networks and traffic flows that often have had limited complexity. However, there has been a recent shift towards using more real-world data in simulations.

\subsection{Algorithms}
\label{sec:problem:alg}
Finally, an RL algorithm must be used to learn a policy. Tabular methods like $Q$-learning and SARSA maintain explicit representations of the $Q$-values for every state and action, which policies directly maximize over given the current state. However, since this is inefficient, deep RL is being increasingly used for RL-based TSC \cite{Wei2019a}. Three broad types of deep RL algorithms exist. \emph{Critic algorithms} use neural networks to approximate $Q$-values; \emph{actor algorithms} use neural networks to directly parameterize policies; and \emph{actor-critic algorithms} combine these ideas by using neural network estimates of $Q$-values to train neural network-based policies \cite{Noaeen2022}. We refer the reader to \cite{Rasheed2020,Haydari2022} for more detailed reviews.

\section{Uncertainty in detection}
\label{sec:detect}

\subsection{Significance of challenges}
\label{sec:detect:prob}
As noted in Section \ref{sec:problem:mdp}, states are described in inputs to RL-based TSC algorithms using abstracted features. These include vehicles' queue lengths, positions, and speeds \cite{Noaeen2022}. Many works take for granted that these features are readily available \cite{Wang2022}. As reported by \cite{Noaeen2022}, 67\% of surveyed papers did not envision any specific data sources. Even in papers where potential data sources were specified, it is unclear how robust the methods would be to detector noise or failure. For instance, among algorithms that use vehicle positions as state features, \cite{Genders2016,VanDerPol2016,Mousavi2017,Liang2019} all used the simulator SUMO to obtain noiseless images of single-intersection toy networks; \cite{Garg2018} extended this approach with a 3D simulator for images from the perspectives of traffic cameras; and \cite{Wei2018} used simulated traffic in SUMO based on flow rates from traffic camera footage. Each of these methods provides a sanitized representation that may not be representative of real-world conditions. Furthermore, the loss of information to noise may cause state aliasing \cite{Spaan2004}, which hinders the generalizability of learned policies to different demand scenarios \cite{Alegre2021}.

\subsection{Lessons from deployments}
\label{sec:detect:deploy}
Types of instruments for traffic sensing include \emph{intrusive detectors} (installed into the road surface) and \emph{non-intrusive detectors} (mounted above the road surface) \cite{Sun2012,Sunkari2019}. Among intrusive detectors, loop detectors are relatively inexpensive, accurate, and robust to weather and time of day, but they are also highly vulnerable to wear and tear \cite{Gibson1998}. When they fail, loop detectors are being increasingly replaced by non-intrusive detectors such as video-based and radar detection systems \cite{Sun2012}, which can be flexibly reconfigured to detect different road segments and vehicle types. However, the accuracy of these systems degrades in inclement weather, and video detectors are also inaccurate at night and on high-speed roads \cite{Sunkari2019,Rhodes2005}. RL-based signal controllers must be designed with these limitations in mind; learning ensembles of models \cite{Lee2021} to capture the strengths of different detectors may improve robustness. Although data about speed and position from connected vehicles can be useful, penetration remains low, so they must be integrated with traditional detector data. \cite{Islam2021} showed in simulations that connected vehicle data could improve adaptive control even with limited penetration. Furthermore, agencies may configure their detectors differently. To account for uncertainty in vehicle stopping positions, for instance, the size of the detection zone behind the stop bar may vary \cite{Emtenan2020}; detectors may also report data at different frequencies \cite{Luyanda2003}. Thus, verifying the mapping from real detector data to abstract state representations is an important task for RL-based TSC.

Agencies often address problems in detection by modifying their detection setup \cite{Sun2012} or by configuring parameters such as passage time (i.e., the amount of time that a phase is extended for upon actuation) \cite{Sunkari2019}. \cite{Smith2013} explicitly addressed error in queue length detection for their adaptive controller SURTRAC. To mitigate underestimation, they used heuristics based on differences in vehicle counts reported by advance and stop bar detectors \cite{Xie2012}. They considered overestimation acceptable, as it provides the algorithm with buffer time; similarly, \cite{Cai2009} found that moderate queue length overestimation significantly improves the performance of adaptive control.

\subsection{Progress toward solutions}
\label{sec:detect:soln}
Two lines of work within RL-based TSC have the potential to address detection uncertainty.

First, various authors have investigated the effects of reducing the dimensionality of the state space. In particular, \cite{Zheng2019a} showed that complex image representations of intersection state achieve inferior performance compared to a simple representation containing only vehicle counts and phases. \cite{Wei2019b} reached similar conclusions with a state representation based on queue length. Both papers also provided optimality results that connected these formulations to traditional methods in TSC. Meanwhile, \cite{Alegre2021,Genders2018} investigated the effects of switching to coarser state representations with a single algorithm. \cite{Genders2018} found that occupancy and speed data (e.g., from loop detectors) yielded near-identical performance to high-fidelity position data (e.g., from cameras). However, the experiments of \cite{Alegre2021} suggested that coarser state discretizations harm generalization across sudden shifts in traffic flow. Regardless, simpler state representations could facilitate identification and debugging of issues caused by detection uncertainty.

Second, other work has attempted to imbue RL-based TSC algorithms with robustness to detection uncertainty. Several methods are analogous to domain randomization in the sim-to-real literature \cite{Andrychowicz2020,Tobin2017}. The approach of \cite{Garg2022} is closest to the sim-to-real literature: they randomize weather and lighting conditions in their traffic simulator and train policies based on the resulting images. \cite{Rodrigues2019} applied Dropout to neural network units to prevent overfitting and thus to learn robust policies. They evaluated their algorithm with a simulation of probabilistic detector failure. As is done in adversarial machine learning, \cite{Tan2020} injected Gaussian noise into queue length observations, and validated their approach with simulations where trucks cause vehicle count overestimation. Meanwhile, to handle miscalibrated measurements, \cite{Wang2022} combined next state prediction with imitation learning from a real traffic controller (SCOOTS), \cite{Li2018a} used autoencoders to denoise input data, and \cite{Aslani2018} evaluated the effects of lane-blocking incidents and detector noise on performance. Finally, in a growing body of work that uses connected vehicle data for RL, \cite{Li2021a} was the first to explicitly address partial observability by adding the phase duration into the state space to learn its indirect impact on delay.

Overall, these methods are helpful approaches for improving the robustness of RL-based TSC to detection uncertainty. However, they should be designed and tuned to address the challenges of specific deployments, leveraging past knowledge to identify and address potential causes of detector noise or failure. It may also help to model partial observability as part of the problem, e.g. by using POMDP-based algorithms.

\section{Reliability of communications}
\label{sec:comm}

\subsection{Significance of challenges}
\label{sec:comm:prob}
As noted in Section \ref{sec:problem:extend}, some level of controller decentralization is often applied in RL-based TSC, because the computational cost of RL may be prohibitive when the state and action space dimensionalities are high. At the same time, to ensure that controllers take the traffic conditions of other intersections into account for signalling decisions, a growing number of works have implemented mechanisms for inter-intersection coordination \cite{Wei2021}. Typical approaches involve sharing states \cite{Ma2020,Chu2020,Xu2020,Wang2021a,Zeng2021,Zhou2019}, actions \cite{Ge2019}, or hidden state representations from neural networks \cite{Wei2019c,Nishi2018} between controllers for neighbouring intersections. While much of this work has focused on designing neural network architectures to leverage shared information (such as graph neural networks \cite{Wang2021a,Zeng2021,Wei2019c,Nishi2018}), less attention has been devoted to the mechanisms by which information must be exchanged in the first place. If there are inconsistencies in the availability of communication infrastructure and detectors between intersections (see also Section \ref{sec:detect}), it is unclear how they may affect the performance of RL-based TSC.

\subsection{Lessons from deployments}
\label{sec:comm:deploy}
In practice, signal controllers are commonly deployed as part of \emph{closed-loop} systems, where control is distributed over three levels. At the top level, \emph{traffic management centres} (TMCs) make policy-based signalling decisions, often involving dialogue with other stakeholders. These decisions are used to configure \emph{field master controllers} (FMCs), which are installed on-site and coordinate multiple \emph{local intersection controllers} (LICs) \cite{Chen2010}. Each FMC aggregates traffic conditions reported by connected LICs to make signalling decisions over a small region; FMCs also synchronize the clocks of LICs to ensure that they are coordinated \cite{Gordon2005,Koonce2008}. As 90\% of TSC systems in the United States are closed-loop \cite{Gettman2007}, upgrades to adaptive control have largely been implemented within this hierarchical organization \cite{Luyanda2003}. LICs may make some limited decisions based on local traffic conditions, but coordination is still largely delegated to FMCs even in adaptive control \cite{Chen2010}. Transitioning to adaptive control has also required agencies to update to Type 2070 or ATC controllers \cite{Gordon2005}, but some controllers in road networks may retain relatively outdated hardware \cite{Koonce2008}. RL-based signal controllers will likely be deployed into such ecosystems, where control is distributed hierarchically and different intersections have different capabilities for control and/or detection. Thus, dec-POMDP formulations and algorithms based on techniques for domain adaptation from the sim-to-real literature may be helpful.

Messages are sent between controllers and TMCs using multiple communication media in modern TSC systems \cite{Gordon2005}. For wired connections, fibre optic cables are increasingly replacing traditional copper wires or coaxial cables. Wireless communication systems implemented using radio or Wi-Fi are also becoming increasingly common \cite{Sun2012}. Thus, communication bandwidth is not likely to be a concern, except in jurisdictions where fibre optic infrastructure is not readily available. However, a major issue reported by agencies in \cite{Sun2012} was connection reliability: poor signal strength often results in data loss or latency. In terms of data formatting, the NTCIP 1202 standard includes standard object definitions for actuated signal controllers, which has also been used for adaptive systems \cite{Gettman2007}. Communications for RL would need to fit into this standard, at least until it is updated (as has already been done for connected vehicles) \cite{Huang2019}. In SURTRAC, \cite{Smith2013} encoded data for communication between neighbouring intersections using JSON messages with standard types.

\subsection{Progress toward solutions}
\label{sec:comm:soln}
One line of work in RL-based TSC has sought to learn more compact representations of information. Although bandwidth is not a concern, reducing message dimensionality could still mitigate the impact of communication failures. Several algorithms directly exchange $Q$-values of learned policies instead of learning from exchanged state representations. In \cite{Wang2016,Liu2017b}, $Q$-values are directly exchanged between neighbours and weighted; \cite{VanDerPol2016,Yang2019,Chu2017} leveraged the max-plus algorithm for coordination graphs, which is known to converge to near-optimality even for cyclic graphs \cite{Kok2005}. Meanwhile, \cite{Xie2020} designed an architecture to exchange information from the previous time step to ensure robustness to latency, and showed that it asymptotically reduces communication relative to neighbour-based approaches by 50\%. \cite{Jiang2022} demonstrated that cumulative rewards can be estimated based only on vehicle counts on inbound approaches.

Some work has also focused on designing RL-based TSC algorithms for hierarchically distributed frameworks of communication and control, which could improve RL's robustness, scalability, and applicability for deployment in closed-loop systems. \cite{Abdoos2021} implemented a two-level architecture where LICs can either act independently or receive joint actions from FMCs based on predictions of the regional traffic state. \cite{Ma2020} introduced a feudal RL algorithm, in which ``manager'' controllers do not directly control the actions of ``worker'' controllers, but instead set goals that influence their rewards. \cite{Xu2021} trained multiple sub-policies that minimize various proxy metrics such as queue length and waiting time, and a high-level controller that adaptively delegates control to sub-policies to minimize the longer-term metric of travel time. However, all of these architectures are conceptual and further work is needed to deploy them.

\section{Compliance and interpretability}
\label{sec:interp}

\subsection{Significance of challenges}
\label{sec:interp:prob}
At the heart of the fact that RL-based TSC algorithms have not been deployed are the potential regulatory and safety risks that are introduced by RL \cite{Wei2019a,Haydari2022}. The issue of trust and safety for RL is by no means exclusive to the domain of TSC \cite{Brunke2022,Garcia2015,Yu2023}, but in this case the stakes are high because controllers must interact with a large number of human users and mistakes may have fatal consequences. For RL-based signal controllers to be trusted, we need to assess --- both prospectively or retrospectively --- whether their decisions comply with standards and reasonable expectations \cite{Ward2020}. However, the proliferation of deep RL algorithms based on complicated state representations runs counter to this goal, as assessment of compliance is not possible if we cannot understand or at least verify their decisions. At the same time, issues of interpretability and safety have rarely been discussed in the literature on RL-based TSC \cite{Noaeen2022} and are more often mentioned as desiderata for future work in reviews \cite{Noaeen2022,Wei2019a,Haydari2022}.

\subsection{Lessons from deployments}
\label{sec:interp:deploy}
In the real world, regulatory frameworks for traffic signalling are often scattershot. In the United States, the federal \textit{Manual on Uniform Traffic Control Devices} \cite{DOT2012} includes standards about the necessity, meaning, and placement of different traffic signals. Many of these standards involve the control of individual movement signals, which would be abstracted away from RL through phase-based action space definitions. However, factors such as yellow change and red clearance intervals are left to ``engineering judgement''. States may impose further requirements on signal timing plans based on regional transportation policies \cite{Koonce2008}. In a review of signal timing policies for 15 states, \cite{Bonneson2009} found recommendations for factors such as minimum green, yellow change, and red clearance intervals, as well as when to serve turn movements. Such recommendations should be incorporated into the design of the RL action space, as was done by \cite{Smith2013} who treated safety constraints as inputs to SURTRAC. Yet, these recommendations can also be arbitrary and dependent on data (e.g., vehicle and pedestrian clearing times \cite{Bonneson2009}), and algorithmic approaches to stakeholder preference learning \cite{Lee2019} may help to find better values.

One common strategy to ensure the safety of signal timing plans is to review common types and causes of crashes in historical data \cite{Bonneson2009}. Naturally, this is a reactive approach that requires crashes to happen in the first place, and crash reports may also be biased by severity or by environmental conditions \cite{Koonce2008}. \emph{Accident modification factors} (AMFs) are a popular method of quantitative analysis; they statistically estimate the effectiveness of particular changes to signal timing plans based on their expected reductions in crash rate \cite{Lord2006,Wu2015,Ma2016}. We are unaware of any work in RL that estimates or uses AMFs, but they may be a valuable pathway to interpretability. The \textit{Highway Safety Manual} also provides standard crash risk assessment models, but these models often require extensive tuning to local conditions \cite{Sun2006,Sun2013,Xie2011}.

\subsection{Progress toward solutions}
\label{sec:interp:soln}
Some work has enhanced the interpretability of RL-based TSC through algorithm design. \cite{Ault2020} focused on learning surrogate policies that are regulatable, i.e. monotonic in state variables, which allows parameters to be viewed as weights. \cite{Jayawardana2021} learned human-auditable decision tree surrogates using VIPER, an algorithm that identifies critical states where suboptimality harms future rewards. Closer to the interpretability literature for machine learning, \cite{Rizzo2019} used SHAP values to analyze how induction loop detections contribute to choices of phases for a controller in a simulated roundabout. They found that advance detectors have higher SHAP values as they are more indicative of congestion. Similarly, \cite{Garg2022} used Grad-CAM to generate heatmaps for image-based inputs. Instead of directly interfacing with the simulator, \cite{Mueller2021} used logical rules based on signal controllers to post-process RL policy outputs for ensuring compliance. 

Further work has applied heuristic modifications to RL algorithms to enforce safety. \cite{Liu2017a} prevented their system from taking actions when pedestrians are detected in crosswalks, and enforced minimum green times for pedestrians. \cite{Essa2020} drew on their models of rear-end conflict rates (based on various observable intersection state features \cite{Essa2018}) to design a reward formulation that minimizes such conflicts. Similarly, \cite{Gong2020} used a binary logistic crash risk model to define crash penalties while also minimizing waiting time. Using a state formulation based on individual signals, \cite{Liao2020} regularized the red light duration of signalling plans to mitigate unsafe behaviour caused by driver frustration with extended red lights. \cite{Yu2020} included yellow change intervals in their action space and added a penalty for emergency braking by vehicles.

While we have reviewed many promising methods that have been developed for the interpretability and safety of RL-based TSC, more work is still needed on determining which of these methods correspond well to stakeholder requirements. Furthermore, there is a substantial literature on \emph{safe reinforcement learning} using constrained optimization \cite{Bohez2019,Ding2020,Liu2022}, which has hitherto not been applied to TSC; it is likely that such work can provide more rigorous theoretical guarantees about algorithm behaviour. We also believe that, to deal with safety failures ethically, work is needed in algorithmic accountability for RL-based signal controllers.

\section{Heterogeneous road users}
\label{sec:roadusers}

\subsection{Significance of challenges}
\label{sec:roadusers:prob}
Traditional models of traffic flow used for TSC assume, simplistically, that all vehicles are identical \cite{Branston1978,Viloria2000}. In reality, the assumption of identical or even unimodal traffic is often unrealistic, because many types of vehicles and road users --- each with different needs and behavioural patterns --- interact with each other on roads. RL algorithms can still implicitly encode these assumptions through simplistic state spaces, since common state variables such as queue length and vehicle position \cite{Wei2019a} do not account for inter-vehicle variation. Although such state formulations can be helpful for deriving optimality results based on traditional models in TSC \cite{Zheng2019a,Wei2019b}, it is unclear how these assumptions may impact the performance and safety of RL-based signal controllers in practice, especially because road users such as pedestrians and cyclists may behave non-intuitively. Dedicated simulators developed for RL-based TSC likewise abstract away inter-vehicle variation \cite{Zhang2019}. \cite{Noaeen2022} found in 160 papers on RL-based TSC that only three accounted for non-private vehicle types, and only one accounted for pedestrians. 

\subsection{Lessons from deployments}
\label{sec:roadusers:deploy}
In practice, agencies make a variety of adjustments to signalling plans to accommodate different classes of road users other than regular passenger vehicles, including pedestrians, cyclists, transit vehicles, and emergency vehicles \cite{Koonce2008}. In this section, we focus on current practice in the field for pedestrians and transit/emergency vehicles. When balancing the needs of different road user classes in RL-based signal controllers, stakeholders' requirements should be taken into account; in the US, for instance, agencies' opinions differ on whether preemption for trains should take priority over pedestrians \cite{Bonneson2009}.

For pedestrians, the simplest option is for the pedestrian signal to be activated in the direction of the through movement, as is implicitly assumed by many works in RL and made explicit in some (e.g., \cite{Guo2019}). However, doing so may cause pedestrians to impede the flow of left-turning and right-turning traffic, which creates safety hazards. In practice, \emph{leading pedestrian intervals} (LePIs) mitigate this risk by allowing pedestrians to start crossing before cars are permitted to make turns \cite{Koonce2008}. Alternative phase sequence designs add lagging pedestrian intervals (after turning phases) or phases exclusively for pedestrians. \cite{Sharma2017} developed a benefit-cost model to assess the safety-delay tradeoffs for LePIs at individual intersections. Beyond safety, additional work has tried to minimize the delay of pedestrians so that they are treated equitably compared to drivers, as codified by regulations in Germany, the UK, and China \cite{Tang2019}. For the deployed SURTRAC system, \cite{Smith2020} adaptively set pedestrian walk intervals based on predicted phase lengths to avoid cutting them short, while \cite{Kothuri2017} considered using vehicular volumes and pedestrian actuation frequencies to switch between controller modes. We are unaware of any work in RL that has explicitly included LePIs as part of the action space formulation. 

As for handling transit and emergency vehicles, typical strategies include the prioritization and preemption of signals. Prioritization handles requests made by vehicles through vehicle-to-infrastructure (V2I) communications, and may or may not result in adjustments to signalling plans. Meanwhile, preemption (often used for firetrucks or trains) deterministically replaces the signal plan with a predefined routine that favours the preempting vehicle. Typically, signal controllers need multiple cycles after preemption to recover from the interruption \cite{Koonce2008}. The adaptive SCATS controller natively implements both prioritization and preemption; compared to prior practice, \cite{Slavin2013} found that SCATS' performance improvements were robust to prioritization, and \cite{Peters2011} found that it could reduce recovery time from preemption. These results suggest the potential of implementing prioritization and preemption with RL-based methods; in particular, explicit modelling of recovery from preemption may further improve recovery times. In addition to interactions at intersections, RL-based signal controllers should also consider the effects of transit and emergency vehicles on traffic between intersections. For instance, when buses are stopped on roads, they may block other traffic from passing. As initial steps towards implementing bus prioritization in the SURTRAC system, \cite{Mahendran2014} delayed the allocation of green time in intersections located downstream from stopped buses, and \cite{Smith2018} predicted bus dwelling times at stops by leveraging V2I communications.

\subsection{Progress toward solutions}
\label{sec:roadusers:soln}
One paper in RL-based TSC was cited by \cite{Noaeen2022} as explicitly modelling pedestrians: \cite{Liu2017a} defined the reward using the weighted average of the local intersection's vehicular queue length, neighbouring intersections' vehicular queue lengths, and the local intersection's pedestrian queue length. Beyond this paper, several other works have explicitly considered pedestrians as part of the problem formulation. \cite{Yin2019} likewise addressed joint vehicle-pedestrian control at intersections, but made no assumptions about pedestrian detector capabilities. \cite{Zhang2021a} used deep RL to control a signalized crosswalk across a road (with the actions being to set the pedestrian signal to green or red), and found that it outperformed actuation under moderate levels of pedestrian demand in simulations. \cite{Aslani2018} analyzed the performance of RL-based TSC in the presence of jaywalking pedestrians that cause vehicles to slow. 

Several works in RL-based TSC have also considered prioritization and preemption. For prioritization, \cite{Garg2022} upweighted buses and emergency vehicles in their throughput-based reward formulation; \cite{Chanloha2014} used a state representation based on the cell transmission traffic model and modelled priority as a binary variable; \cite{Shabestray2019} adopted an implicit approach based on minimizing delay per person instead of per vehicle; \cite{Zhang2017} and \cite{Guo2021} both considered prioritization for trams, with the former's rewards being based on tram schedule adherence and the latter using model predictive control to model driver behaviour; and \cite{Kumar2021} adaptively altered vehicles' priorities depending on queue length, waiting time, and emergency vehicle presence. For preemption, \cite{Su2022} learned TSC policies for emergency vehicle routing with rewards that encourage low vehicle density, and \cite{Su2021} used RL to learn policies for notifying connected vehicles to clear out lanes for emergency vehicles to pass.

Lastly, \cite{Mueller2021} included demand data from the field for multiple types of road users --- including pedestrians, cyclists, motorcyclists, trucks, and buses --- in their benchmark simulation for RL-based TSC, LemgoRL, which is based on a real road network; they also included pedestrian waiting times in rewards and enforced minimum pedestrian green times. There is a need to connect high-fidelity simulations such as LemgoRL to the various approaches for handling different road user classes that we outlined above, so as to ensure their ecological validity.

\section{Conclusion}
\label{sec:conclusion}
We have reviewed four barriers to the deployment of RL-based controllers for TSC. Each of these barriers has been insufficiently addressed by the majority of new work in RL-based TSC, which has focused on algorithmic contributions. However, TSC algorithms do not exist in a vacuum --- they must be trained based on data from detectors, interface with signals through controllers, and control the movements of a variety of road users. Challenges both intrinsic to RL algorithms and in other pipeline components may cascade into failures with significant implications for the efficiency and safety of transportation infrastructure. Based on our literature review, we suggested ways in which further work in RL-based TSC could address these challenges.

\begin{itemize}
    \item \textbf{Uncertainty in detection}. Instead of assuming that state features are available without noise, considerations about detectors should be part of the algorithm design process. State spaces should be designed based on the detector types that were used to collect the data; more complex representations are not necessarily more useful. Techniques in ensemble learning and robustness for RL could deal with noise and failure, especially if multiple detector modalities are available.
    
    \item \textbf{Reliability of communications}. RL algorithms will likely be deployed in closed-loop signal control systems with hierarchical control architectures, where domain adaptation may be necessary. Learning concise message representations may be useful for inter-intersection communication, but it is more important to consider potential failures in communication between controllers and TMCs due to poor signal strength than restrictions on communication bandwidth.
    
    \item \textbf{Compliance and interpretability}. Flexibility to enforce different types of constraints, possibly through action and reward formulations, is necessary to ensure that RL algorithms remain compliant as traffic conditions evolve. Models for deep RL must be designed so that their output policies are easily interpretable and auditable by relevant stakeholders, and the evaluation of policies should be based not just on performance metrics but also on safety (e.g., using AMFs).
    
    \item \textbf{Heterogeneous road users}. Although simple simulations with uniform vehicles can increase RL training efficiency, the resulting policies may be suboptimal in practice. Action space formulations should be designed to incorporate signal plan modifications for different types of road users, including LePIs, prioritization, and preemption. The most important goal is to achieve equity in minimizing the delay and maximizing the safety of different road user classes.
\end{itemize}

Echoing the recommendations of \cite{Perrault2019}, we emphasize the importance of engaging in consultation with agency stakeholders and experts in TSC for RL practitioners. This can break down information silos that would otherwise prevent the recognition of issues during requirements engineering and integration (cf. \cite{Nahar2022}); we could not have identified these challenges ourselves without engaging with the literature on traditional TSC. Additionally, as we discussed, the practicalities of these challenges --- including the availability and configuration of detectors, signalling constraints, and the priorities of different road users --- will often vary depending on the statuses of road networks and their responsible agencies. While benchmark simulations based on synthetic networks facilitate evaluation, we advocate for the creation of more simulations like \cite{Mueller2021} that incorporate realistic domain constraints. RL algorithms that are trained using such benchmarks would likely have better generalizability and robustness in deployments.

More generally, we uncovered a diversity of work that addresses each challenge, which previous reviews of TSC have not comprehensively surveyed. This suggests that RL-based TSC is closer to deployment than might be suggested by a review of state-of-the-art methods. If future developments focus on combining algorithmic improvements with both real-world considerations and reproducibility techniques to facilitate collaboration \cite{Pineau2021}, we believe that the integration of RL to improve real-world transportation infrastructure is within reach.

\begin{acknowledgments}
The authors thank Christian K\"astner, Eunsuk Kang, Stephanie Milani, Peide Huang, Ryan Shi, and Steven Jecmen for useful information and suggestions that they provided to support the drafting of this review. This work was supported in part by a research grant from Mobility21, NSF grant IIS-2046640 (CAREER), and the Tang Family Endowed Innovation Fund.
\end{acknowledgments}

\bibliography{ref}
\end{document}